\def\year{2022}\relax
\documentclass[letterpaper]{article} % DO NOT CHANGE THIS
\usepackage{aaai22}  % DO NOT CHANGE THIS
\usepackage{times}  % DO NOT CHANGE THIS
\usepackage{helvet}  % DO NOT CHANGE THIS
\usepackage{courier}  % DO NOT CHANGE THIS
\usepackage[hyphens]{url}  % DO NOT CHANGE THIS
\usepackage{graphicx} % DO NOT CHANGE THIS
\usepackage{natbib}  % DO NOT CHANGE THIS AND DO NOT ADD ANY OPTIONS TO IT
\usepackage{caption} % DO NOT CHANGE THIS AND DO NOT ADD ANY OPTIONS TO IT
\DeclareCaptionStyle{ruled}{labelfont=normalfont,labelsep=colon,strut=off} % DO NOT CHANGE THIS
\usepackage{gensymb}  % For degree symbol
\usepackage{algorithm}
\usepackage[noend]{algpseudocode} % The noend option will discard end of loop lines like end if, end for, end procedure, etc.

\usepackage[dvipsnames]{xcolor}

\usepackage{newfloat}
\usepackage{listings}
\floatstyle{ruled}
\newfloat{listing}{tb}{lst}{}
\floatname{listing}{Listing}
\nocopyright
\usepackage{hhline} % RVAdded
\usepackage{amsmath}
\usepackage{multirow}
\usepackage{placeins}
\usepackage{amsthm}
\usepackage{amssymb}

\title{Non-Blocking Batch A* (Technical Report)}
\author{
    Rishi Veerapaneni, Maxim Likhachev
}
\affiliations{
    Robotics Institute, Carnegie Mellon University \\
    \{rveerapa, mlikhach\}@andrew.cmu.edu
}

\begin{document}

\maketitle

\begin{abstract}
Heuristic search has traditionally relied on hand-crafted or programmatically derived heuristics. Neural networks (NNs) are newer powerful tools which can be used to learn complex mappings from states to cost-to-go heuristics. However, their slow single inference time is a large overhead that can substantially slow down planning time in optimized heuristic search implementations. Several recent works have described ways to take advantage of NN's batch computations to decrease overhead in planning, while retaining bounds on (sub)optimality. However, all these methods have used the NN heuristic in a ``blocking" manner while building up their batches, and have ignored possible fast-to-compute admissible heuristics (e.g. existing classically derived heuristics) that are usually available to use. We introduce Non-Blocking Batch A* (NBBA*), a bounded suboptimal method which lazily computes the NN heuristic in batches while allowing expansions informed by a non-NN heuristic. We show how this subtle but important change can lead to substantial reductions in expansions compared to the current blocking alternative, and see that the performance is related to the information difference between the batch computed NN and fast non-NN heuristic.
\end{abstract}

\section*{Introduction}\label{sec:intro}
Heuristic search is a common deterministic motion planning class of algorithms that plans paths from start and goal positions by searching over different paths. Heuristic search algorithms use heuristics, along with other mechanisms, to speed up the path planning process.

Modern machine learning techniques (e.g. reinforcement, imitation and supervised learning) learn complex mappings from states to values that can be viewed as cost-to-go heuristics. Recent work have started to use these techniques to learn neural network heuristics on combinatorial search and pathfinding problems which are better able to guide search compared to existing heuristics \cite{deepCubeA2019, deepCubeAQ2021, learningHeuristicA2020, learningGlobalHF2011, scaleFree2021}.

A major bottleneck of using neural networks (NNs) during search is that the computational overhead to compute a single heuristic value is significantly higher than non-neural network approaches. Neural network inferences are known to be faster when computed on batches of inputs allowing them to take advantage of their single instruction multiple data (SIMD) nature and specialized SIMD hardware (i.e. GPUs) rather than computed many times individually. 

Recently, there have been several works that describe different ways for Batch A*. \citet{deepCubeA2019} describes ``Batch Weighted A*", BWAS, which expands K states from their OPEN queue. This unfortunately does not maintain any suboptimality bounds with the inadmissible learnt heuristics. \citet{kfocal} introduces ``K-Focal Search" which improves upon BWAS by utilizing a fast admissible heuristic ($h_{fast}$) and a $w_{so}$ bounded suboptimal Focal Search and expanding K nodes at once from the FOCAL queue that have f-values $\leq \min_n g(n) + h_{fast}(n)$. The admissible $h_{fast}$ (which is usually a classically derived heuristic, or the zero heuristic if none are available) maintains bounded suboptimality in the Focal queue and enables batch computation of the inadmissible $h_{NN}$ during each batch expansion. Their method shows large runtime benefits by increasing the batch size B to decrease the runtime overhead of $h_{NN}$. Another recent work \citet{optimalSearchNN} describes a batch A* mechanism that maintains optimality, but contains several restrictions on its use case as it requires an admissible learnt heuristic and a problem where many states may share the same heuristic values.

Our method, Non-Blocking Batch A*, is closest to \citet{kfocal} as we utilize a $w_{so}$ bounded suboptimal search, an admissible fast heuristic $h_{fast}$, and an inadmissible neural network heuristic $h_{NN}$. Our two key differences are that instead of expanding K states at once, we lazily compute the NN heuristic in batches and that we eliminate waiting by allowing the search to expand states without requiring the NN heuristic to be computed. We show that these are required to sustain reduction in nodes generated and obtain speed-ups.

\section*{Non-Blocking Batch A*}
``K-Focal Search" is constructed using a NN function $h_{succ}()$ that takes in a state $n$ and returns the $h$ values of all its successors $n'$. Thus, expanding K nodes at once allowed them to compute the heuristics of all their successors at once and insert them into queues accordingly. This requirement on the NN function is restrictive as the same NN cannot work with different action spaces as the successor states and corresponding heuristic values would change when using different action spaces. Additionally this requirement generally increases the complexity of the learning problem as we must map one parent state to many successor heuristic values. We instead formulate our work with a NN heuristic function $h_{state}$ which returns the heuristic value of one state which simplifies the learning problem and is independent of the action space. Note we can easily construct $h_{succ}()$ by using this single-state NN $h_{state}$ internally.

The K-Focal Search analogue using $h_{state}$ is to expand K nodes, collect their successor states into a single large batch B, and pass the batch into $h_{state}$ to generate the heuristic values. However, this does not maintain a fixed batch size B as expanded states can have different numbers of valid successors. An iterative approach can mitigate this issue by expanding states sequentially, adding successor states onto a waitlist until a target batchsize of B is approximately reached, and then calling $h_{state}$ and inserting the nodes into the queues.

We show that this approach has two fundamental problem; 1) the waitlist blocks ``deeper" expansions, and 2) the search only uses the non-neural network fast admissible heuristic ($h_{fast}$) to maintain a lower bound and never utilizes it to guide search. Specifically, once a node $n$ is expanded, its successors are placed in the waitlist until the waitlist is full, preventing any successor from being expanded in the meanwhile. This prevents rapid expansions of subsequent states, essentially forcing us away from a greedy heuristic search and into a K-breadth greedy search. This construction also ignores the fact that we may have a fairly informative $h_{fast}$ that we can use along with $h_{state}$, instead of solely using $h_{state}$ for guiding search efforts. Additionally, finding a solution path of length $L$ (not cost) will now require a minimum of $L$ batches which can be a large problem if B and L are large.

Our alternative removes this blocking behaviour by utilizing $h_{fast}$ along with $h_{state}$, allowing our search algorithm to expand consecutive nodes guided by $h_{fast}$ without needing to wait for the expensive batch computation.

Algorithm \ref{alg:nbba} shows the full pseudo-code of our method. The main changes are commented in blue. Specifically, line 18 inserts nodes into FOCAL based on $h_{fast}$ instead of solely residing in the waitlist as in ``K-Focal Search". Lines 19-22 call $h_{NN}$ on a batch of states once the batchsize has surpassed $B$, and updates $h_{fast}$ to use the these values (as they can be cached into $h_{fast}$ and accessed quickly). Lines 7-9 check if $h_{fast}$ has a new updated value (i.e. from an earlier $h_{NN}$ batch call) and updates $n$'s priority and reinserts accordingly.

\begin{algorithm}[tb]
\caption{Non-Blocking Batch A*}
\label{alg:nbba}
\textbf{Hyper-parameters}: Suboptimality bound $w_{so}$, Heuristic weight $w_h$ \\ 
\textbf{Input}: $n_{start}$, atGoal(), Batch NN heuristic $h_{NN}()$, Fast admissible heuristic $h_{fast}()$ \\
\textbf{Output}: Path from $n_{start}$ with suboptimality $\leq w_{so}$
\begin{algorithmic}[1] %[1] enables line numbers
\State OPEN = FOCAL = $\{n_{start}\}$, Waitlist = $\{\}$
\While{FOCAL $\neq \emptyset $}
    \State $n \gets$ FOCAL.pop()
    \State OPEN.remove($n$)
    \If{atGoal($n$)}
        \State \textbf{return} Solution backtracking from $n$
    \EndIf
    \Statex \textcolor{blue}{\Comment{Lazily update $h_{focal}$ and reinsert if needed}}
    \If{$n.h_{focal} \neq h_{fast}(n)$ }
        \State $n.h_{focal} \gets h_{fast}(n)$
        \State AddToQueues($n$)
        % \State OPEN.insert($n$)
    \Else \textcolor{black}{\Comment{Generate successors}}
    \For{$n'$ $\in$ succ($n$)}
        % \Statex \Comment{Always insert n to open}
        \State $n'.g \gets n.g + cost(n,n')$
        \State $n'.h_{open}, n.h_{focal} \gets h_{fast}(n')$
        \State AddToQueues($n$)
    \EndFor
    \EndIf
    
    \Statex \textcolor{black}{\Comment{Update FOCAL}}
    % \State $LB \gets \max(LB,n.F_{open})$
    \ForAll {$n' \in OPEN, \notin FOCAL $} 
        \If {$n'.F_{open} \leq w_{so}*n.F_{open}$}
            \State Waitlist.insert($n'$)
            \State \textcolor{blue}{FOCAL.insert($n'$) \Comment{Non-blocking}}
        \EndIf
    \EndFor
    
    % \Statex \textcolor{blue}{\Comment{Compute batch heuristic and update $h_{fast}$}}
    \If {Waitlist.size() $\geq B$}
        \Statex \textcolor{blue}{\Comment{Compute batch heuristic and update $h_{fast}$}}
        \State $h_{1:K} \gets h_{NN}(n_i \in \text{Waitlist})$
        \State $\forall {n_i \in \text{Waitlist}}, h_{fast}(n_i) \gets h_{i} $
        % \State Update $n.F$ values for $n \in \text{OPEN, FOCAL}$
        \State Waitlist $= \{\}$
    \EndIf
\EndWhile
\State \textbf{return} No solution
\Statex
\Procedure{AddToQueues}{$n$}:
    \State $n.F_{open} \gets n.g + n.h_{open}$
    \State $n.F_{focal} \gets n.g + w_{h}*n.h_{focal}$
    \State OPEN.insert($n$)
    \If{$n.F_{open} \leq w_{so} * \min_{n'} n'.F_{open}$}
        \State FOCAL.insert($n$)
    \EndIf
\EndProcedure
\end{algorithmic}
\end{algorithm}

\begin{figure}[ht]
    \centering
    \includegraphics[width=0.45\textwidth]{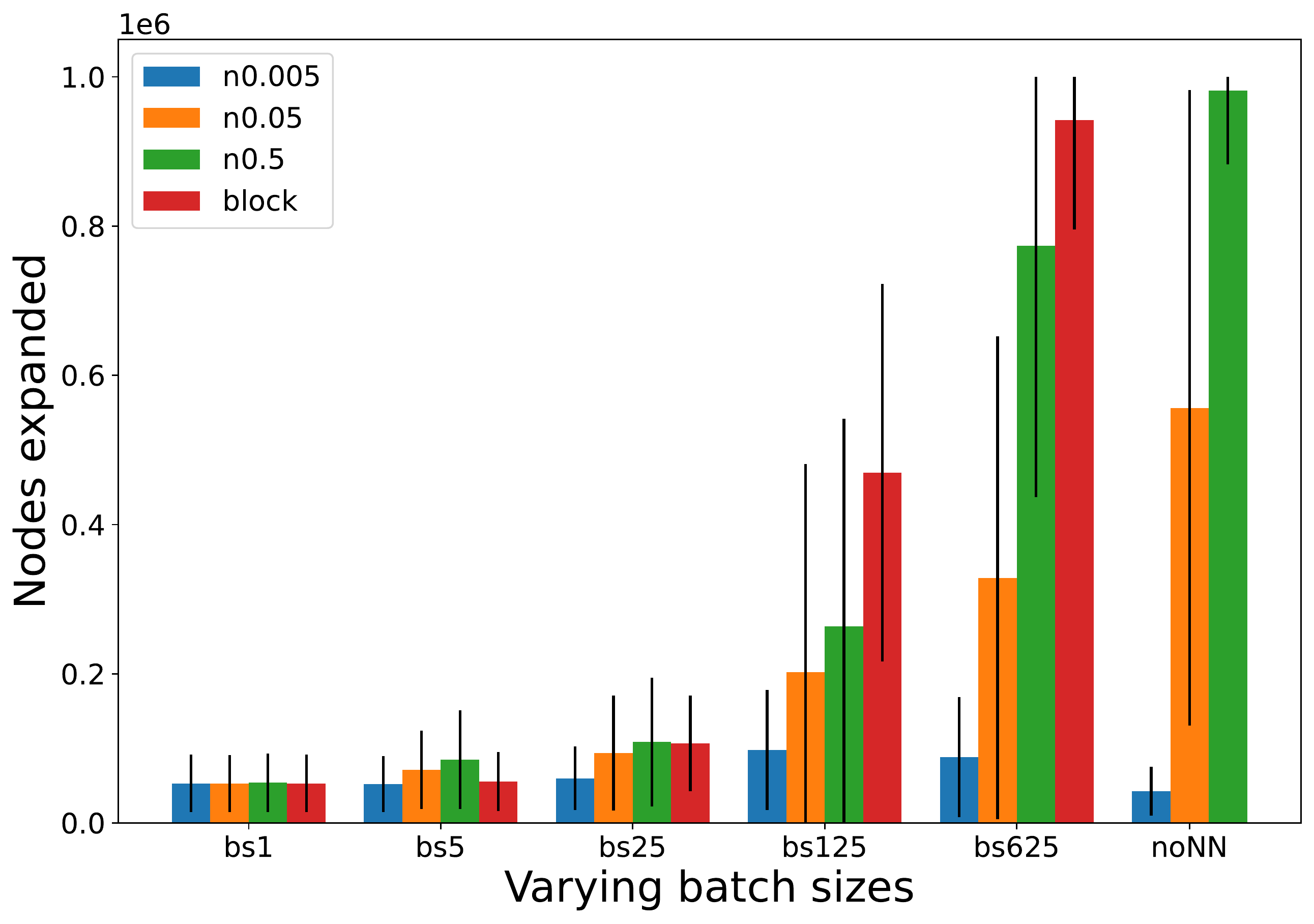}
    \caption{\footnotesize \textbf{Node expansions |} We see the numbers of nodes expanded during search by our method with various fidelity $h_{fast}$ (blue, orange, red) and compare it against the blocking version (red) across different batch sizes. Note that the blocking version does not depend on $h_{fast}$ in FOCAL so varying $h_{fast}$ did not change performance much. The last column shows the results of just running the noisy $h_{fast}$; we see from the orange and green baselines struggle. We see that increasing batch size increases the number of expansions as predicted, and that our non-blocking version generally outperforms the blocking version.}
    \label{fig:figure-expansion}
    \vspace*{-2mm}
\end{figure}

\section*{NBBA* Experimental Results}
\subsection{Set-up}
We show results on path-finding in a 2D Map using a Neural Network. Our planning space is $(x,y,\theta)$ with the following three actions: 1) forward in respect to the $\theta$ heading, 2) stationary turn left 90$\degree$, 3) turn right stationary turn left 90$\degree$. Our goals are $(x_g, y_g)$ tuples and our point mass agent starts at positions $(x_s, y_s, 0)$. We test on 30 instances on a large 8192x8192 binary ``sand-trap" obstacle map (5\% obstacles randomly generated) where all transitions in the map are valid and unit cost except for transitions exiting sand-traps which cost 100. Figures show the mean and standard deviation across the 30 instances. We run methods with $w_{so}=w_h=2.5$ until they find the goal or expand a maximum of 1 million nodes. This set-up was arbitrarily used from an existing project, but from spot-checking briefly on different parameters the trends from these experiments seem to hold across different 2D grid world scenarios (and likely to other domains as well).

Our method requires two heuristics, the base fast heuristic $h_{fast}$ and the batch NN heuristic $h_{NN}$. To relate the difference between the quality of the heuristics provided by $h_{fast}$ and $h_{NN}$, we make both $h_{NN}$ noisy versions of the Manhattan distance of $(x,y)$. Concretely, let $h_m(s)$ by the true Manhattan distance heuristic. We define a ``noisy" version with $k \in [0,1]$ as of $h_{m_k} = h_m*N$ where $N \sim U[1-k, 1]$. A noisy $h_{m_k}$ with $k = 0$ is a ``strong" heuristic, perfectly noiseless and equal to $h_m$, while increase $k$ makes $h_{m_k}$ increasingly weak and noisy. Note that although $h_{m_0}$ has a clear meaning, $h_{m_1}$ does not and is not the same as a perfectly uninformative heuristic.

Since our goal is to demonstrate the effectiveness of our non-blocking nature, we do not train a neural network to output $h_m$. Instead we use a dummy $h_{NN}$ which creates a random tensor input of size $(B,242)$ and passes it into a small 3-layer neural network for a timing purposes, but actually returns a noisy $h_{m_{0.01}}$. We compare our method against K-Focal which uses $h_{fast}$ for OPEN queue and $h_{NN}$ in FOCAL, and implement the iterative batch building variant for a clear comparison against our method. We label this K-Focal variant as ``blocking" in our experiments.

\subsection{Results}
Figure \ref{fig:figure-expansion} shows the number of nodes expanded during search by our method as well as the blocking version with varying $h_{fast}$ and batch sizes. The right most column shows the results of running search with just $h_{fast}$ where we see that a noisy $n$ level of $0.005$ runs fast (blue), $n = 0.05$ struggles but still manages to find paths with an average of 500k expansions (orange), while $n=0.5$ is not informative and results in a majority of failures.
We first observe that in general, increasing batch sizes increases the number of nodes expanded, which makes sense as increasing the batch size requires expanding more nodes until $h_{NN}$ is called again and the updated heuristic values can be used. However we observe that our non-blocking version outperforms the blocking version across most instances. If $h_{fast}$ gets stronger, the performance gap increases as our non-blocking algorithm allows successive expansions guided by $h_{fast}$.

\begin{figure}[ht!]
    \centering
    \includegraphics[width=0.45\textwidth]{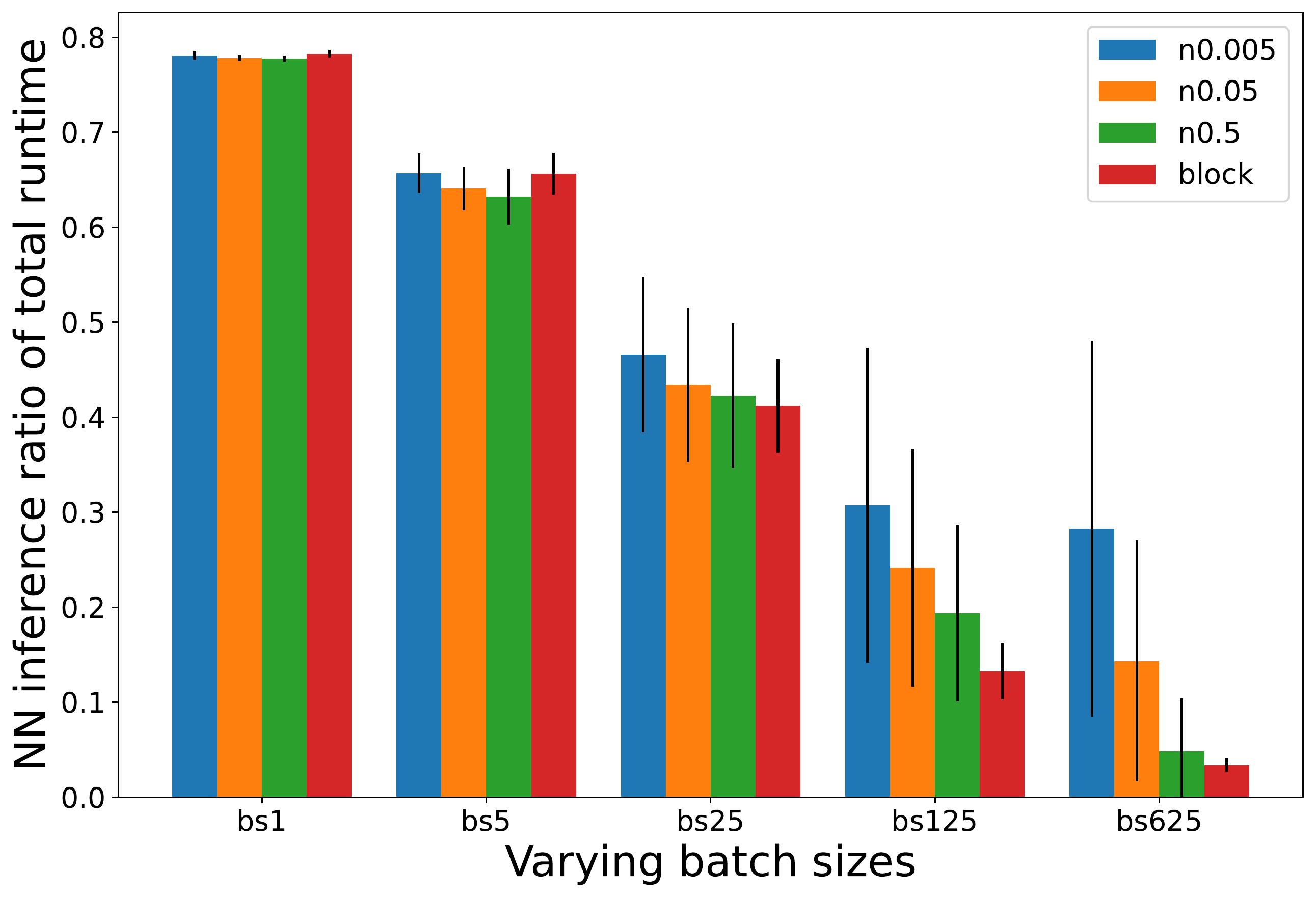}
    \caption{\footnotesize \textbf{Neural network overhead |} We plot the total inference time for computing the heuristic compared to the total runtime. We first observe that increasing batchsize reduces runtime as expected. The exact ratio changes based on the strength of $h_{fast}$. As $h_{fast}$ becomes stronger, each $h_{NN}$ call becomes less useful and more of a relative overhead.}
    \label{fig:figure-ratio}
    \vspace*{-2mm}
\end{figure}

\begin{figure}[ht!]
    \centering
    \includegraphics[width=0.45\textwidth]{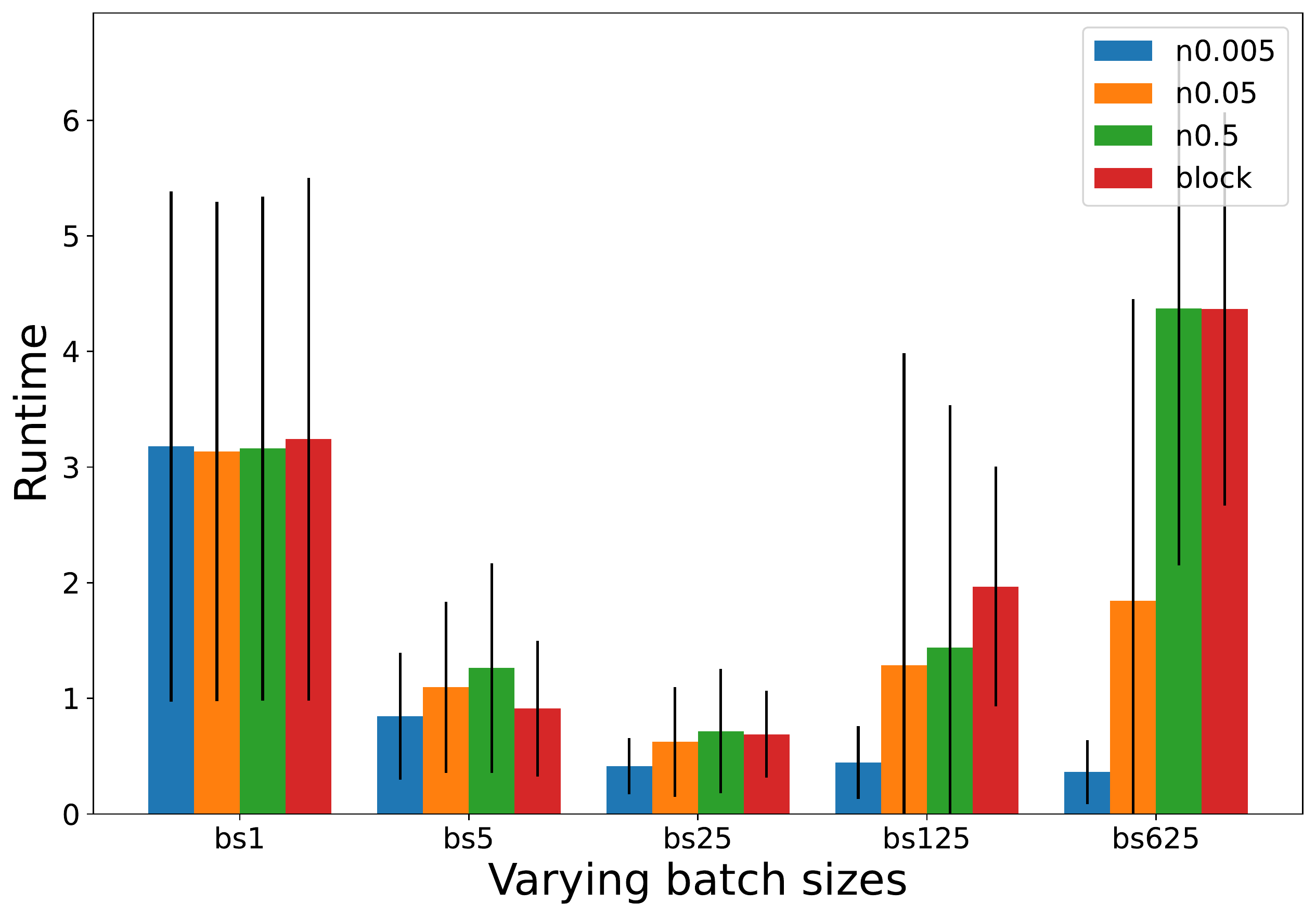}
    \caption{\footnotesize \textbf{Runtime |} We compare the runtime of blocking versus non-blocking methods. We see that for larger batchsizes (e.g. 25, 125), our non-blocking variant offers runtime benefits at the cost of increased variance due to the use of $h_{fast}$. Note that both the green and red methods reach the maximum node expansion and terminate early on the rightmost side (batchsize of 625), but that there our orange non-blocking method performs much faster even though the orange heuristic itself is pretty weak (see Figure \ref{fig:figure-expansion}).}
    \label{fig:figure-runtime}
    \vspace*{-2mm}
\end{figure}

Figure \ref{fig:figure-ratio} highlights the ratio of inference time that $h_{NN}$ takes compared to the overall search time. Increasing batch size decreases the ratio as expected. However, we see that the absolute value depends on the strength of $h_{fast}$. We are unclear why this exactly occurs but can offer some initial thoughts. As $h_{fast}$ becomes stronger, the search advances deeper between $h_{NN}$ calls, generating more unique children and populating the waitlist faster (which we confirmed experimentally, although it does not explain all the difference). Each $h_{NN}$ call occurs relatively more frequent, but at the same time adds marginally less information than $h_{fast}$. We feel this ``marginally less" information has a side-effect effecting the runtime ratio but are unable to find concrete cause and effect with our initial effort.

Finally, we show the aggregate runtime of NBBA* and the blocking K-Focal. At a low batch size and higher $h_{fast}$ noise, the blocking behaviour can be beneficial as it eliminates expanding extra nodes and reduces overall runtime. However with larger batchsizes this blocking behaviour becomes a bottleneck resulting in our method being faster on average.

% We hypothesize that this occurs as $h_{NN}$ relative use decreases as $h_{fast}$ becomes stronger. Thus with $n = 0.005$, the search is guided strongly by $h_{fast}$ and $h_{NN}$ expansions result in a large overhead for relatively little gain. Thus the ratio of number of $h_{NN}$ expansions to number of total nodes expanded is low. On the otherhand, with $n = 0.5$, $h_{fast}$ is barely informative, so the number of $h_{NN}$ to total node expansions will be higher.

\section*{Conclusion}
We show a general method for incorporating a learned heuristic $h_{NN}$ in bounded suboptimal manner that leverages batch computation as well as the fast but not as informative base heuristic $h_{fast}$. Non-Blocking Batch A* (NBBA*) lazily creates batches for $h_{NN}$ while permitting expansions based on $h_{fast}$. We show this can provide speed ups over the current state of the art K-Focal Search depending on the strength of $h_{fast}$ and batch size.
This technical report has several limitations; namely that we do not conduct robust experimentation across different domains and neural networks sizes. Further analysis could also be done into determining how the strength of $h_{fast}$ changes the ratio of time spent on $h_{NN}$'s inference time as mentioned in the previous section. However we believe and hope that NBBA* will be useful to future practitioners incorporating learned heuristics with search.
\textbf{Acknowledgements: } Rishi V. would like to thank Shivam S. for their helpful discussion in proofreading this report.

\clearpage
\bibliography{ref} 

\end{document}